\pdfoutput=1

\documentclass[11pt]{article}

\usepackage[]{acl}

\usepackage{times}
\usepackage{latexsym}

\usepackage[T1]{fontenc}

\usepackage[utf8]{inputenc}

\usepackage{microtype}

\usepackage{inconsolata}

\usepackage{times}
\usepackage{latexsym}
\usepackage{amsfonts}
\usepackage{amsmath}
\usepackage{subcaption}
\usepackage{graphicx}

\usepackage{xcolor}
\usepackage{CJKutf8}

\usepackage{tabularx}
\usepackage{booktabs}
\usepackage{multirow}

\usepackage{enumitem}

\newcommand{\zhsmall}[1]{\begin{CJK*}{UTF8}{gbsn}\small{#1}\end{CJK*}}

\title{Teaching Large Language Models an Unseen Language on the Fly}

\author{Chen Zhang,\ \ Xiao Liu,\ \  Jiuheng Lin,\ \ Yansong Feng\thanks{Corresponding author.} \\
Peking University \\
{\tt \{zhangch,lxlisa,fengyansong\}@pku.edu.cn} \\
{\tt linjiuheng@stu.pku.edu.cn}\\
}
\begin{document}
\maketitle
\begin{abstract}
Existing large language models struggle to support numerous low-resource languages, particularly the extremely low-resource ones, for which there is minimal training data available for effective parameter updating. 
We thus investigate whether LLMs can learn a new language on the fly solely through prompting. 
To study this question, we collect a research suite for Zhuang, a language supported by no LLMs currently. 
We introduce \textsc{DiPMT++}, a framework for adapting LLMs to unseen languages by in-context learning. 
Using a dictionary and 5K parallel sentences only, \textsc{DiPMT++} significantly enhances the performance of GPT-4 from 0 to 16 BLEU for Chinese-to-Zhuang translation and achieves 32 BLEU for Zhuang-to-Chinese translation.
We also validate the effectiveness of our framework on Kalamang, another unseen language.
Furthermore, we demonstrate the practical utility of \textsc{DiPMT++} in aiding humans in translating completely unseen languages, which could contribute to the preservation of linguistic diversity.
\end{abstract}

\section{Introduction}
Existing large language models (LLMs) provide robust support for many high-resource languages, but their support for numerous low-resource languages is limited~\cite{ahuja-etal-2023-mega}. 
To adapt LLMs to low-resource languages, continual pre-training or adaptors are commonly employed~\cite{pfeiffer-etal-2020-mad,yong-etal-2023-bloom}. However, a corpus of only a few thousand sentences is insufficient to update the model parameters effectively for learning extremely low-resource languages~\cite{joshi-etal-2020-state}. 
Considering the inductive and mimicking capabilities of LLMs, an interesting research question arises: Can LLMs learn a new low-resource language on the fly solely through prompting? 
This learning paradigm could enable more efficient utilization of limited resources and holds significant potential in the preservation and education of underrepresented languages.

\begin{figure}[t]
\centering
\includegraphics[scale=0.5]{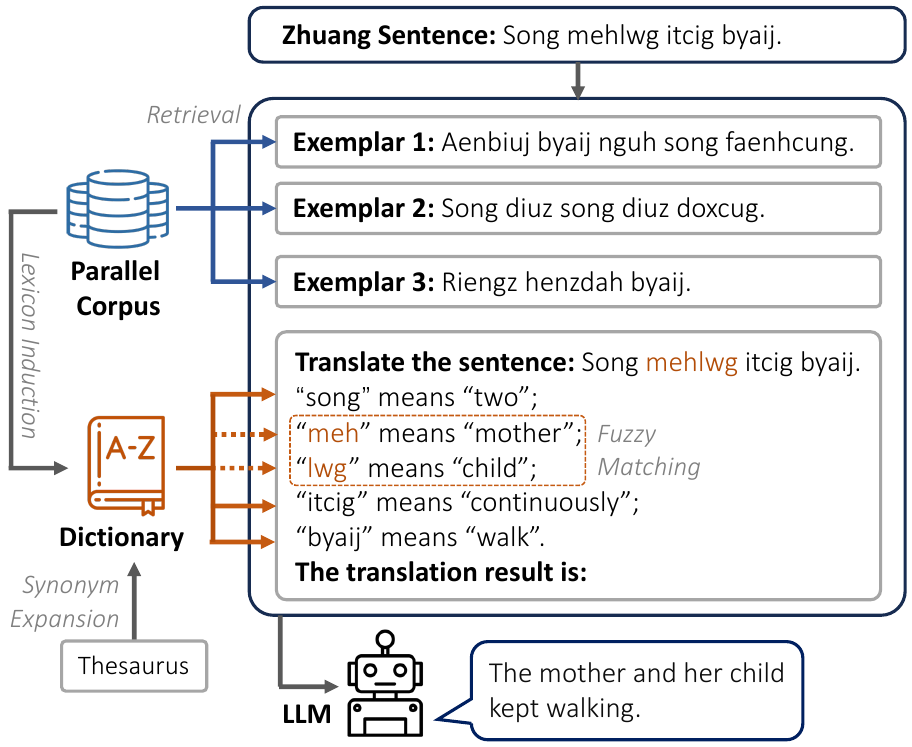}
\caption{An example of translating a Zhuang sentence with \textsc{DiPMT++}.}
\label{fig:method}
\end{figure}

To explore this question, we choose Zhuang (ISO 639-1: za; ISO 639-3: zha), an extremely low-resource language, as our focus. 
It is the language spoken by the Zhuang people of Southern China.\footnote{See more information about Zhuang in Appendix \ref{appendix:zhuang}.} 
There are no open-source natural language processing (NLP) datasets in Zhuang, and existing LLMs do not support this language.
Therefore, we curate \textsc{ZhuangBench}, a research suite for Zhuang, comprising a dictionary, a parallel corpus of 5K Zhuang-Chinese sentences, and a machine translation test set.
\textsc{ZhuangBench} is not only a valuable linguistic resource for this extremely low-resource language but also a challenging benchmark for LLMs, with which we can investigate how models learn an entirely new language.

We focus on the machine translation (MT) task in \textsc{ZhuangBench}. 
Previous research such as \textsc{DiPMT}~\cite{ghazvininejad2023dictionary} has explored translation in low-resource languages via prompting, providing translations for rare words with existing dictionaries.
This method is tailored for languages where LLMs already possess fundamental capabilities, especially basic knowledge for syntax, which might be learned in the form of $n$-gram language modeling.
However, when confronted with an entirely new language where LLMs need to acquire its vocabulary and grammar from scratch, previous prompting methods hardly work.

In this work, we introduce \textsc{DiPMT++}, a framework to efficiently adapt LLMs to an unseen language via in-context learning (ICL). 
Built upon \textsc{DiPMT}, our method provides models with the meanings of words appearing in the source sentence, as illustrated in Figure~\ref{fig:method}. Given the inherent incompleteness of the dictionary for low-resource languages, we enhance the lexical coverage by revisiting traditional techniques like bilingual lexicon induction and synonym expansion.
To aid models in grasping basic syntax, we retrieve closely related exemplars from parallel corpora to construct the ICL prompt.

We evaluate \textsc{DiPMT++} with various models as backbones on \textsc{ZhuangBench}.
\textsc{DiPMT++} consistently outperforms other prompting baselines and smaller models finetuned for the task.
Specifically, when paired with GPT-4, which initially exhibits near-zero performance, \textsc{DiPMT++} achieves impressive BLEU scores of 15.7 for Chinese-to-Zhuang translation and 31.9 for Zhuang-to-Chinese translation.
We additionally evaluate \textsc{DiPMT++} on MTOB~\cite{tanzer2024a}, a benchmark for translation between English and Kalamang, another low-resource language unseen to LLMs. Our framework achieves the best performance in most settings, exhibiting its language-agnostic effectiveness.
Going beyond \textsc{DiPMT++}, we also explore more potential strategies to teach LLMs a new language's syntax through prompts.

To investigate the applicability of \textsc{DiPMT++} in realistic scenarios, we conduct a user study of unseen language translation. We recruit participants who have no knowledge of Zhuang, and ask them to conduct translation with the given linguistic sources. Experiments reveal that if we assist humans with \textsc{DiPMT++}, their translation improves in both quality and efficiency. 
This shows our framework's great potential in preserving endangered languages.

Our contributions are as follows:

$\bullet$ We present \textsc{ZhuangBench}, a challenging benchmark for LLMs to translate an unseen language with limited linguistic resources.

$\bullet$ We develop an ICL framework \textsc{DiPMT++} for on-the-fly language learning, which has proven effective on two benchmarks, and explore more strategies for enhancing LLMs' acquisition of lexical and syntactic knowledge.

$\bullet$  We showcase that \textsc{DiPMT++} can assist humans in translating unseen languages, which could benefit the preservation of linguistic diversity.

Our code and data are available to the public\footnote{\url{https://github.com/luciusssss/ZhuangBench}}.

\section{Related Works}

\paragraph{Adapting LLMs to Low-Resource Languages}
Continual pretraining on monolingual texts is a common practice to adapt LLMs to low-resource languages~\cite{yong-etal-2023-bloom,zhang2023mc}. 
Techniques such as MAD-X~\cite{pfeiffer-etal-2020-mad,pfeiffer-etal-2021-unks} and LoRA~\cite{hu2021lora} are used to improve the training efficiency.
\citet{yang2023bigtrans} design a high- to low-resource curriculum for multilingual continual pertaining. 
\citet{purkayastha2023romanization} attempt to romanize the unseen scripts for language adaptation.

Another line of work directly adapts instruction-tuned LLMs through supervised fine-tuning (SFT) by constructing cross-lingual instructions~\cite{cahyawijaya2023instructalign} or leveraging pivot languages~\cite{zhang2023plug}.
\citet{yong-etal-2023-bloom} find that SFT is sometimes more efficient for language adaptation than continual pretraining.

Unlike previous works, we explore the possibility of on-the-fly language adaptation through prompting given minimal linguistic resources.

\paragraph{Machine Translation with LLMs}
Recent works demonstrate the effectiveness of prompting an LLM with a few MT examples~\cite{DBLP:conf/icml/0006HB23,vilar-etal-2023-prompting,garcia2023unreasonable}.
Following them, researchers explore different strategies for ICL exemplar selection, from the perspectives of term recall~\cite{agrawal-etal-2023-context}, knowledge relevance~\cite{he2023exploring}, and cultural awareness~\cite{yao2023empowering}. However, these strategies are primarily tailored for high-resource languages.

\citet{robinson-etal-2023-chatgpt} show that LLMs still fail to translate low-resource languages properly.
Existing methods designed for low-resource languages include incorporating dictionaries~\cite{ghazvininejad2023dictionary,elsner-needle-2023-translating}, using high-resource languages as pivots~\cite{jiao2023chatgpt}, and refining results from smaller MT systems~\cite{cheng2023scale}.
In contrast to them, our work tackles an even more challenging scenario: translating languages entirely unseen by LLMs.

\section{Dataset: \textsc{ZhuangBench}}
We present \textsc{ZhuangBench}, the first NLP research suite for Zhuang, consisting of a Zhuang-Chinese dictionary, 
a Zhuang-Chinese parallel corpus, and a Zhuang-Chinese translation test set.
It can be used for various NLP tasks, such as word sense disambiguation, cross-lingual retrieval, and machine translation.
Here, we especially focus on the task of performing Zhuang-Chinese translation using the dictionary and parallel corpus.

\paragraph{Dictionary}
The Zhuang-Chinese dictionary is collected from an online dictionary site\footnote{\url{https://zha_zho.en-academic.com/}}, with 16,031 Zhuang words.
The average number of senses for each word is 1.4. 
We also convert it to a Chinese-Zhuang dictionary with 13,618 Chinese words, with an average of 2.2 translations per word.

\paragraph{Parallel Corpus}
The parallel corpus contains 4,944 Zhuang-Chinese sentence pairs from multiple sources. 
2,135 pairs are obtained from the Chinese and Zhuang versions of the Government Work Reports in China, where we map each Chinese sentence to its corresponding Zhuang translation. 
2,127 pairs are collected from Zhuang textbooks.
The remaining 682 pairs are example sentences in the dictionary.
See more statistics in Appendix~\ref{appendix:dataset}.

\paragraph{Translation Test Set}
We provide a hold-out test set with 200 sentence pairs for evaluation. 
It includes instances of three difficulty levels: 75 \textit{easy} instances, 60 \textit{medium} ones, and 65 \textit{hard} ones.

The \textit{easy} subsets are composed of sentences from an elementary textbook.
Other subsets are collected from the official Zhuang Language Proficiency Test (Vahcuengh Sawcuengh Suijbingz Gaujsi, V.S.S.G.) in China, which is divided into three levels: elementary, intermediate, and advanced. 
The translation instances from the elementary and intermediate levels of V.S.S.G. form our \textit{medium} subset while those from the advanced level constitute our \textit{hard} subset. 
In Appendix~\ref{appendix:dataset}, we list statistics and examples for each level. The three subsets of different difficulties exhibit a gradual increase in vocabulary coverage and sentence complexity.

\section{Method}
We introduce \textsc{DiPMT++}, a language-agnostic framework to adapt LLMs to an unseen language efficiently. 
It can serve as a strong baseline for the machine translation task in \textsc{ZhuangBench}.

\subsection{Preliminary: \textsc{DiPMT}}
\textsc{DiPMT}~\cite{ghazvininejad2023dictionary} is a prompting-based method for low-resource language translation. Given a source sentence, the model looks up in the dictionary for the meaning of rare words and adds them to the prompt directly with the format \textit{in this context, the word ``[source word]'' means ``[target word]''}. 
The prompt is adopted in an in-context learning manner, with $k$ demonstrations before the current testing instance.

\textsc{DiPMT} is designed for languages that current models perform moderately well (10 - 30 BLEU scores). The authors claim that \textsc{DiPMT} is not suitable for languages where current models perform poorly (< 10 BLEU points), as they assume that \textit{the performance for those is too low to expect reasonable translations even when incorporating external information}.

\subsection{Our Method: \textsc{DiPMT++}}
\label{subsec:method}
One may ask what we can do to help LLMs understand those extremely low-resource languages, for which we only have a parallel corpus of a few thousand sentences.
Contrary to the pessimistic view of \citet{ghazvininejad2023dictionary}, we hypothesize that it is possible to teach LLMs a new language solely through prompting, considering LLMs' impressive ability to infer and mimic. 
We make extensions to the \textsc{DiPMT} framework so that it can be applied to extremely low-resource languages. 
Leveraging the powerful reasoning capabilities of LLMs, we propose \textsc{DiPMT++}, a new method that allows LLMs to understand a completely new language with minimal resources.

Following \textsc{DiPMT}, we cast the on-the-fly machine translation as an ICL task and incorporate knowledge from bilingual dictionaries.
\textsc{DiPMT++} makes two key modifications to \textsc{DiPMT} as follows. 
See the prompt template in Appendix~\ref{appendix:implementation_details}.

\paragraph{Improved Lexical Coverage}
\textsc{DiPMT} only provides meanings for less frequent words in the sentence. For an unseen language, we need to provide translations for as many words in the sentence as possible to the model. However, not all words can be found in the dictionary for a low-resource language. 
\textsc{DiPMT++} attempts to improve lexical coverage of the prompt by revisiting traditional statistical methods and linguistic resources. Specifically, we use the following three strategies: 

$\bullet$ \textbf{Fuzzy Matching:} Due to various morphological transformations such as derivation, inflection, compounding, etc., words appearing in a sentence may not be directly found in the dictionary. However, for a low-resource language, we lack available morphological analysis tools. In this case, string matching algorithms such as forward/backward maximum matching can quickly find potentially relevant entries from the dictionary.

$\bullet$ \textbf{Bilingual Lexicon Induction:} Even a small-scale parallel corpus might contain words not included in the dictionary. Traditional statistical methods such as GIZA++~\cite{och-ney-2003-systematic} can efficiently mine bilingual lexicon from the corpus, which could complement the dictionary.

$\bullet$ \textbf{Synonym Expansion:}
When translating a word from a high-resource language, it is not always possible to find a direct translation in the dictionary. However, the dictionary may contain entries for synonyms of the word. This problem can be alleviated by expanding the dictionary to include a list of synonyms~\cite{shi2005synonym}.

\paragraph{Syntactically-Informed Exemplar}
In \textsc{DiPMT}, the exemplars are fixed and have limited relevance to the testing instance. Their function is only to demonstrate the task. 
It hardly works when the LLMs have little knowledge about the grammar of an unseen language.
\textsc{DiPMT++} attempts to dynamically select exemplars with higher relevance and encourage models to infer elementary syntactic information from the exemplars. 
For a testing instance, its exemplars are retrieved from a parallel corpus. 
Although there are advanced retrievers such as DPR~\cite{karpukhin-etal-2020-dense}, they require additional training and do not support extremely low-resource languages like Zhuang. 
Therefore, we apply BM25~\cite{robertson2009probabilistic}, a language-agnostic retrieval algorithm, for exemplar retrieval.

\begin{table*}[t]
\setlength\tabcolsep{2.8pt}
\centering
\begin{small}
\begin{tabular}{l|rrrr|rrrr}
\toprule
\multirow{2}{*}{\textbf{Model}}  & \multicolumn{4}{c|}{\textbf{Chinese $\to$ Zhuang (BLEU / chrF)}} &  \multicolumn{4}{c}{\textbf{Zhuang $\to$ Chinese (BLEU / chrF)}} \\
 & \multicolumn{1}{c}{\textbf{\textit{easy}}} & \multicolumn{1}{c}{\textbf{\textit{medium}}} & \multicolumn{1}{c}{\textbf{\textit{hard}}} & \multicolumn{1}{c|}{\textbf{All}} & \multicolumn{1}{c}{\textbf{\textit{easy}}} & \multicolumn{1}{c}{\textbf{\textit{medium}}} & \multicolumn{1}{c}{\textbf{\textit{hard}}} & \multicolumn{1}{c}{\textbf{All}} \\
\midrule
\multicolumn{9}{c}{Baselines} \\
\midrule
mT5-base (Finetune) & 0.1 /\ \ \  7.5 & 0.2 /\ \ \  7.9 & 0.0 /\ \ \  7.9 & 0.1 /\ \ \  7.8 &  0.2 /\ \ \  1.2 & 0.2 /\ \ \  1.4 & 0.2 /\ \ \  1.3 & 0.2 /\ \ \  1.3 \\
mT5-large (Finetune) & 3.3 / 21.0 & 0.8 / 15.0 & 0.3 / 12.5 & 1.1 / 15.1 & 2.5 /\ \ \  4.6 & 0.4 /\ \ \  2.1 & 1.1 /\ \ \  1.9 & 1.3 /\ \ \  2.6 \\
Llama-2-7B-chat (Finetune) & 29.0 / 54.7 & 5.9 / 34.2 & 1.0 / 25.5 & 6.5 / 33.2 & 21.2 / 19.4 & 10.1 / 11.2 & 5.8 / 7.6 & 11.3 / 11.2 \\
\midrule
GPT-3.5 (Direct) & 0.2 / 14.0 & 0.3 / 15.8 & 0.5 / 17.2 & 0.3 / 16.1 & 0.6 /\ \ \  3.8 & 0.5 /\ \ \  3.2 & 0.2 /\ \ \  3.4 & 0.3 /\ \ \  3.5 \\
GPT-4 (Direct) & 0.1 /\ \ \   6.2 & 0.1 /\ \ \  7.2 & 0.1 /\ \ \  7.2 & 0.0 /\ \ \  7.0 & 0.9 /\ \ \  3.9 & 0.8 /\ \ \  3.7 & 1.7 /\ \ \  3.9 & 1.4 /\ \ \  3.9 \\
\midrule
Qwen-7B-chat (\textsc{DiPMT}) & 3.9 / 28.2 & 1.0 / 24.4 & 0.7 / 23.3 & 1.8 / 24.8 & 12.8 / 13.9 & 10.2 / 11.2 & 3.3 /\ \ \  5.7 & 8.3 /\ \ \  9.3 \\
Qwen-14B-chat (\textsc{DiPMT}) & 8.4 / 35.4 & 4.2 / 30.8 & 2.1 / 26.1 & 4.4 / 29.6 & 20.6 / 18.1 & 17.0 / 16.0 & 5.9 /\ \ \  7.7 & 12.7 / 12.8 \\
Qwen-72B-chat (\textsc{DiPMT}) & 9.2 / 36.6 & 4.5 / 31.7 & 3.1 / 29.4 & 5.1 / 31.7 & 23.0 / 21.3 & 19.8 / 17.9 & 9.8 / 10.4 & 16.3 / 15.1 \\
\midrule
\multicolumn{9}{c}{\textsc{DiPMT++} (Ours)} \\
\midrule
Llama-2-7B-chat & 14.8 / 47.4 & 4.4 / 33.5 & 1.1 / 28.0 & 5.9 / 33.9 & 19.8 / 19.3 & 9.3 / 10.9 & 4.3 / 6.5 & 9.7 / 10.7 \\
Llama-2-13B-chat & 21.5 / 51.6 & 7.1 / 37.7 & 3.0 / 32.4 & 9.0 / 38.2 & 22.8 / 20.6 & 9.1 / 11.5 & 6.0 / 7.5 &  10.7 / 11.6 \\
Llama-2-70B-chat & 21.5 / 50.7 & 8.2 / 41.3 & 2.4 / 34.6 & 9.3 / 40.6 & 26.4 / 25.4 & 12.5 / 13.9 & 6.2 / 8.7 & 12.7 / 13.8 \\
\midrule
Qwen-7B-chat  & 17.6 / 48.4 & 5.1 / 37.0 & 2.8 / 31.5 &  7.6 / 36.9 & 22.4 / 25.2 & 11.7 / 15.5 & 5.5 / 8.5 & 11.4 / 14.3 \\
Qwen-14B-chat & 28.2 / 55.8 & 10.6 / 42.5 & 4.8 / 34.9 & 12.6 / 41.7 & 34.3 / 31.0 & 21.6 / 20.8 & 9.1 / 9.7 & 19.5 / 17.8   \\
Qwen-72B-chat & \textbf{31.1} / \textbf{58.1} & \underline{13.9} / \underline{43.4} & \textbf{9.9} / \underline{40.4} & \textbf{16.4} / \underline{45.1} & \underline{43.6} / \underline{39.4} & \underline{30.1} / \underline{28.0} & \underline{18.7} / \underline{19.9} & \underline{27.3} / \underline{26.4} \\
\midrule

GPT-3.5 & 25.7 / 53.8 & 11.3 / 42.6 & \underline{7.6} / 39.6 & 13.3 / 43.5 & 34.3 / 31.8 & 17.1 / 18.0 & 16.3 / 17.5 & 20.1 / 20.5 \\

GPT-4 & \underline{30.7} / \underline{57.3} & \textbf{15.1} / \textbf{45.4} & 7.4 / \textbf{41.7} & \underline{15.7} / \textbf{46.1} & \textbf{48.3} / \textbf{43.2} & \textbf{35.0} / \textbf{31.0} & \textbf{22.8} / \textbf{21.8} &  \textbf{31.9} / \textbf{29.1} \\
\bottomrule
\end{tabular}
\end{small}
\caption{Performance of different methods on the test set of \textsc{ZhuangBench}. We use 3-shot exemplars for prompting-based methods. The best scores are made \textbf{bold}, with the second \underline{underlined}.}
\label{tab:main_experiment}
\end{table*}

\section{Experiments }
\subsection{Experimental Setup}

\paragraph{Backbone Models}
We use three types of models as the backbone of \textbf{DiPMT++}: (1) \textbf{Llama-2-chat}~\cite{touvron2023Llama}, an open-source English-centric model, (2) \textbf{Qwen-chat}~\cite{bai2023qwen}, a bilingual model for English and Chinese, and (3) \textbf{GPT-3.5} and \textbf{GPT-4}~\cite{openai2023gpt4}, two commercial multilingual models\footnote{The versions of the OpenAI APIs are \texttt{gpt-3.5-turbo-0125} and \texttt{gpt-4-0125-preview}.}.

\paragraph{Baselines}
We adopt a variety of baselines for comparison. 
(1) \textbf{Finetune}: Finetuning models with parallel sentences. 
(2) \textbf{Direct}: Directly asking LLMs to perform translations without providing ICL exemplars, which, to some extent, reflects whether the LLM already knows the language. 
(3) \textbf{\textsc{DiPMT}}~\cite{ghazvininejad2023dictionary}: Using the original design of \textsc{DiPMT} for LLM prompting.

\paragraph{Metrics}
We use BLEU~\cite{papineni-etal-2002-bleu} and chrF~\cite{popovic-2015-chrf}, implemented by sacreBLEU~\cite{post-2018-call}. 
BLEU is a word-level metric while chrF focuses on the character level.

\paragraph{Tokenization}
All the models used in our experiments support Chinese. 
Because Zhuang adopts a Latin script, these models can tokenize Zhuang texts into subwords, or characters at least, without producing UNK. For example, Llama-2's tokenizer tokenizes \textit{Liuz coengmingz.} into \texttt{[`\_Li', `uz', `\_co', `eng', `ming', `z', `.']}.

See more implementation details in Appendix~\ref{appendix:implementation_details}.

\subsection{Results and Analyses}

In Table~\ref{tab:main_experiment}, we report the results on the Chinese-to-Zhuang (zh2za) and Zhuang-to-Chinese (za2zh) translation task of \textsc{ZhuangBench}. 
See samples of output from different models in Appendix~\ref{appendix:case_study}.

\paragraph{Finetuning vs. Prompting}
Finetuning pretrained models is a common practice for low-resource machine translations~\cite{adelani-etal-2022-thousand}.
Fintuned smaller models like mT5 still have BLEU scores close to zero for Zhuang\footnote{Although mT5-large achieves 15.1 chrF on zh2za, the model outputs are almost non-sense, as shown by the example in Appendix~\ref{appendix:case_study}. As Zhuang uses a Latin script with 26 characters, even a meaningless sequence would likely have a non-zero chrF score.}.
Through finetuning, Llama-2-7B-chat can develop a basic understanding of Zhuang. It performs particularly well on the \textit{easy} subset of zh2za translation but still struggles with more challenging instances. 
We refrain from finetuning larger models due to the substantial computational resources required.

Compared to the high expense of finetuning, prompting with \textsc{DiPMT++} requires no training while delivering comparable or even superior performance when combined with larger models.
Prompting Llama-2-7B-chat with \textsc{DiPMT++} only lags finetuning by 0.6 BLEU on zh2za and by 1.6 BLEU on za2zh. 
Despite having little prior knowledge about Zhuang, as evidenced by their poor performance in direct prompting, GPT-3.5 and GPT-4 achieve excellent results on the translation tasks of varying difficulty levels with the assistance of \textsc{DiPMT++}. 
For instance, GPT-4 paired with \textsc{DiPMT++} achieves a remarkable BLEU score of 31.9 on za2zh, which might be qualified for practical use.

\paragraph{\textsc{DiPMT} vs. \textsc{DiPMT++}}
We compare \textsc{DiPMT} and \textsc{DiPMT++} with Qwen-chat as the backbone model.
Although useful for mid-source languages, the original \textsc{DiPMT} has limited ability to assist LLMs in understanding a completely new language. Even Qwen-72B-chat, the largest version of Qwen-chat, achieves only 5.1 BLEU on Chinese-to-Zhuang translation. 
After introducing two simple extensions, \textsc{DiPMT++} activate the reasoning ability of LLMs and greatly boost the performance. 
For example, \textsc{DiPMT++} increases the BLEU scores of Qwen-72B-chat by 122\% on zh2za and by 67\% on za2zh over the original \textsc{DiPMT}.
We will further discuss how each design in \textsc{DiPMT++} contributes to the overall performance in Section~\ref{sec:dicussion}.

\paragraph{Model Scale}
Regarding model scales, we observe that the performance steadily improves with the increase of model parameters for Llama-2 and Qwen.
Since Qwen has a better Chinese capability than the English-centric Llama-2, a 14B Qwen model can outperform a 70B Llama-2.
GPT-4 outperforms all other models, demonstrating its excellent reasoning ability. 
It is worth noting that the open-source Qwen-72B-chat performs comparably to the closed-source GPT-4 on the zh2za task, which is an encouraging result for more transparent and reproducible research on low-resource NLP. 

\subsection{Additional Experiments on Other Languages}
\paragraph{Another Unseen Language: Kalamang}
It is extremely hard to identify a language completely unseen by current LLMs and collect enough resources for it.
Besides \textsc{ZhuangBench}, the only suitable evaluation dataset is MTOB, from a contemporary work~\cite{tanzer2024a}. It consists of translation tasks between English (eng) and Kalamang (kgv), another low-resource language unseen by current LLMs.

We report preliminary results on \textsc{MTOB} in Table~\ref{tab:kalamang}.
\textsc{DiPMT++} outperforms the baseline in the original paper of \textsc{MTOB} across most settings. This further proves that \textsc{DiPMT++} is a language-agnostic framework and can adapt to different low-resource languages without extra effort. 
See details in Appendix~\ref{appendix:kalamang}.

\begin{table}[t]
\centering
\begin{small}
\begin{tabular}{l|cc|cc}
\toprule
& \multicolumn{2}{c|}{\textbf{eng2kgv}} & \multicolumn{2}{c}{\textbf{kgv2eng}} \\
& \textbf{BLEU} & \textbf{chrF} & \textbf{BLEU} & \textbf{chrF} \\
\midrule
\multicolumn{5}{c}{Original Baseline \cite{tanzer2024a}} \\
\midrule
Llama-2-13B & 0.0 & 28.8 & ~~4.8 & 29.8 \\
Llama-2-70B & 0.0 & \textbf{40.1} & ~~9.4 & 34.9 \\
GPT-3.5 & 0.0 & 30.6 & ~~6.6 & 31.1 \\
\midrule
\multicolumn{5}{c}{\textsc{DiPMT++} (Ours)} \\
\midrule
Llama-2-13B & \underline{3.7} & 31.5 & 11.3 & 35.2\\
Llama-2-70B & \textbf{4.4} & \underline{35.7} & \textbf{12.3} & \underline{36.3} \\
GPT-3.5 & 2.9 & 35.5 & \underline{11.4} & \textbf{37.0} \\
\bottomrule
\end{tabular}
\end{small}
\caption{Results of the original baseline  from \citet{tanzer2024a} and \textsc{DiPMT++} on the test set of \textsc{MTOB}. The original baseline is the \textbf{W} + \textbf{S} setting in \citet{tanzer2024a}. The best scores are made \textbf{bold}, with the second \underline{underlined}.}
\label{tab:kalamang}
\end{table}

\paragraph{Seen Languages}
We also evaluate \textsc{DiPMT++} on 7 low-resource languages that might have been seen during pre-training. Unlike unseen languages such as Zhuang and Kalamang, LLMs can translate these languages with a non-zero BLEU score by zero-shot prompting. 
\textsc{DiPMT++} can still improve the translation quality for extremely low-resource languages, whose BLEU scores are below 10 originally.
See details in Appendix~\ref{appendix:seen_lang}.

\section{Discussion}
\label{sec:dicussion}
Here we delve into two important questions when adapting LLMs to an unseen language. One is how to improve the coverage of lexical explanations given limited or incomplete resources. We analyze the three strategies used in \textsc{DiPMT++} to alleviate the out-of-dictionary problem. 
The other is how to teach LLMs syntactic rules solely through prompting.
We go beyond \textsc{DiPMT++} and investigate more potential strategies to help LLMs learn syntax implicitly or explicitly.

\subsection{Improving Lexical Coverage}
During the translation, one might not find all the words in the dictionary.
For example, with the original dictionary in \textsc{ZhuangBench}, we can only find the entries for 67\% of the Zhuang words and 47\% of the Chinese words in the test set of \textsc{ZhuangBench}.
As described in Section~\ref{subsec:method}, we adopt three strategies to improve the lexical coverage of the prompt.
Here we analyze how they contribute to the performance of \textsc{DiPMT++}.

\textbf{The introduction of these strategies significantly improves the lexical coverage in the prompt.} 
By running GIZA++~\cite{och-ney-2003-systematic}, an effective algorithm for mining bilingual lexicon, on the parallel corpus of \textsc{ZhuangBench}, we obtain 2,051 new Chinese-Zhuang word pairs. Adding them to the dictionary helps increase the lexical coverage on the test set of \textsc{ZhuangBench}, from 67\% to 79\% for Zhuang words and from 47\% to 53\% for Chinese words.
By incorporating a Chinese synonym list, we add 18,491 new entries to the Chinese-Zhuang dictionary, which further increases the lexical coverage of the Chinese words to 66\%.
For the resting uncovered Zhuang or Chinese words, we search in the dictionary with forward/backward maximal matching and provide the top two potentially related words.

\textbf{The increase in lexical coverage is propagated to the improvement in the translation quality.} 
Table~\ref{tab:ablation_study} shows ablation studies for the three strategies with Qwen-14B-chat.
All three strategies contribute to the overall translation performance.  
For zh2za translation, the scores drop most after we remove the fuzzy matching, since this strategy helps provide lexical information for up to 44\% of the Chinese words.
Meanwhile, bilingual lexicon induction is the most important for za2zh translation, as it introduces the highest number of gold word-level translations for the testing instances.

\textbf{Different strategies can address different types of out-of-dictionary problems.} 
We qualitatively analyze the types of words recalled by each strategy on \textsc{ZhuangBench}.
Fuzzy matching greatly helps address the compound words in Chinese and Zhuang. For example, the out-of-dictionary Chinese word \zhsmall{日常生活} (daily life) is decomposed into \zhsmall{日常} (daily) and \zhsmall{生活} (life) by forward maximal matching. Similarly, the Zhuang word \textit{mehlwg} (mother and child) is decomposed into \textit{meh} (mother) and \textit{lwg} (child).
By bilingual lexicon induction on our corpus, we can mine common words overlooked by the dictionaries, i.e., \textit{soujgih} $\Leftrightarrow$ \zhsmall{手机} (mobile phone), or newly-invented words, i.e., \textit{lienhcanjnieb} $\Leftrightarrow$ \zhsmall{产业链} (industry chain).
Since these strategies are language-agnostic, they can also be applied to other languages and alleviate the out-of-dictionary problems caused by certain morphological phenomena.

\begin{table}[t]
\centering
\begin{small}
\begin{tabular}{l|cc|cc}
\toprule
& \multicolumn{2}{c|}{\textbf{zh2za}} & \multicolumn{2}{c}{\textbf{za2zh}} \\
& \textbf{BLEU} & \textbf{chrF} & \textbf{BLEU} & \textbf{chrF} \\
\midrule
\textsc{DiPMT++} & \textbf{12.6} & \textbf{41.7} & \textbf{19.5} & \textbf{17.8} \\
\midrule
w/o Fuzzy & 9.1 & 35.3 & 18.9 & 17.4 \\
w/o BLI & 10.1 & 36.5 & 12.2 & 12.0 \\
w/o Synonym & 11.7 & 38.7 & 19.5 & 17.8  \\
\bottomrule
\end{tabular}
\end{small}
\caption{Ablation study of the three strategies for improving lexical coverage with Qwen-14B-chat on \textsc{ZhuangBench}. BLI is short for Bilingual Lexicon Induction. Note that there is no synonym list available for Zhuang, so the w/o Synonym setting of za2zh has the same scores with \textsc{DiPMT++}.}
\label{tab:ablation_study}
\end{table}

\subsection{Learning Syntax}
For extremely low-resource languages like Zhuang, there is neither a large-scale monolingual corpus available for language modeling nor well-defined formal grammar such as context-free grammar (CFG) ready to use.
Given the strong reasoning ability of LLMs, one might wonder to what extent LLMs can learn the syntax of an unseen language, either \textit{implicitly} or \textit{explicitly}, through prompting.
We conduct a preliminary study of different strategies for teaching LLMs syntax and reveal what strategies work and what might not currently.

\subsubsection{Implicit Learning}
Based on the imitation ability of LLMs, we expect them to learn the syntax explicitly from the given texts in the unseen language. 
One possible approach is retrieving similar sentences for reference, as \textsc{DiPMT++} does. Another is providing a piece of monolingual texts in the prompt to familiarize the LLM with this new language.

\paragraph{Learn from \textsc{DiPMT++} Exemplars}
In \textsc{DiPMT++}, we retrieve exemplars from the corpus with BM25, a language-agnostic retrieval algorithm, hoping LLMs can infer shallow syntax from them. 
Besides BM25, we try two other strategies. 
One is randomly sampling exemplars. 
The other is retrieval based on part-of-speech (POS) sequence\footnote{As there are no POS taggers available for Zhuang, the POS sequence of a Zhuang sentence is approximated by concatenating the POS tag of each word's Chinese translation.}, assuming that sentences with similar POS sequences (measured by Levenshtein distance) may share similar syntactical structures.

As shown in Table~\ref{tab:learn_from_exemplar}, the exemplars retrieved by BM25 greatly improve the translation quality over the random sampling, as they contain richer lexical and syntactic information related to the testing instance.
POS-based retrieval might not provide much assistance to the model's syntactic learning, as the abstraction from natural language sentences to POS sequences is overly simplified and the POS sequences are sometimes noisy.

\begin{table}[t]
\centering
\begin{small}
\begin{tabular}{l|cc|cc}
\toprule
& \multicolumn{2}{c|}{\textbf{zh2za}} & \multicolumn{2}{c}{\textbf{za2zh}} \\
& \textbf{BLEU} & \textbf{chrF} & \textbf{BLEU} & \textbf{chrF} \\
\midrule
Random & ~~9.2 & 40.0 & 14.7 & 14.1 \\
POS & ~~7.8 & 30.1 & 12.1 & 12.8 \\
BM25 & \textbf{12.6} & \textbf{41.7} & \textbf{19.5} & \textbf{17.8} \\
\bottomrule
\end{tabular}
\end{small}
\caption{Results of using different strategies for obtaining exemplars in \textsc{DiPMT++}, using Qwen-14B-chat on \textsc{ZhuangBench}.}
\label{tab:learn_from_exemplar}
\end{table}

\paragraph{Learn from Monolingual Texts}
Inspired by learning language modeling from a large-scale corpus, we wonder whether LLMs can quickly familiarize themselves with the syntax of an unseen language through a small piece of monolingual texts in the prompt when we have only a few thousand tokens of texts for a low-resource language.
Therefore, we add monolingual Zhuang texts to the \textsc{DiPMT++} prompt\footnote{In this study, we did not design sophisticated strategies for the selection of monolingual texts. We directly extract monolingual texts of varying lengths from a news story in Zhuang, instead of retrieving texts for different testing instances. 
In the future, we will collect more monolingual corpora and explore clever strategies for selecting proper monolingual texts.}, hoping that it can help LLMs generate more coherent Zhuang sentences.

We test this strategy with GPT-3.5\footnote{We initially test this strategy using Qwen-14B-chat. This model fails to produce meaningful outputs when presented with an additional 1K tokens of Zhuang texts in the prompt. 
Incorporating large portions of texts in an unseen language might overwhelm this medium-sized LLM.}.
Table~\ref{tab:add_monolingual} shows a 0.9 BLEU gain in zh2za translation after adding 1K Zhuang tokens. Notably, the \textit{easy} subset sees a substantial 2.8-point increase. 
As the length of the monolingual text increases, we do not observe a continuous increase in performance. 
We find that when provided with 5K monolingual text, the model becomes overconfident in this language and fabricates more words not present in the prompt.
We refrain from adding longer monolingual texts than 5K for the high expense.

\begin{table}[t]
\centering
\begin{small}
\begin{tabular}{l|ccc|c}
\toprule
& \textbf{\textit{easy}} & \textbf{\textit{medium}} & \textbf{\textit{hard}} & \textbf{All} \\
\midrule
\textsc{DiPMT++} & 25.7 & 11.3 & 7.6  & 13.3 \\
\midrule
+1K Tokens & \textbf{28.5}  & 11.3  & \textbf{8.2} & \textbf{14.2}  \\
+2K Tokens & 26.9  & \textbf{13.1}  & 6.4  & 13.6  \\
+5K Tokens & 26.5  & 11.6  & 6.2  & 13.1  \\
\bottomrule
\end{tabular}
\end{small}
\caption{BLEU scores of adding different lengths of monolingual texts to the prompt of \textsc{DiPMT++}, using GPT-3.5 on the zh2za task of \textsc{ZhuangBench}.}
\label{tab:add_monolingual}
\end{table}

\subsubsection{Explicit Learning}
Besides allowing LLMs to infer the syntax of an unseen language, one may directly feed them with specific syntactic rules.
However, we might not have a ready-to-use collection of grammatical rules for low-resource languages. Moreover, adding every rule to a prompt is not feasible in practice.
Thus, we focus on a particular syntactical phenomenon and investigate whether LLMs can comprehend it when provided with explicit rules.

We choose the order of modifiers and modified elements as our research focus.
Different from Chinese and English, Zhuang usually places modifiers after modified elements. 
We select 10 zh2za testing samples where GPT-3.5 fails on the order of modifiers and modified elements, and try different strategies to incorporate this syntactic rule into the prompt. 

First, we attempt to directly declare a rule of word order between modifiers and modified elements in the prompt, i.e., \textit{different from Chinese, Zhuang puts the modifiers after the modified elements}.
GPT-3.5 correctly addresses only 1 of the 10 testing examples.

Second, we use chain-of-thought (CoT) reasoning to encourage the model to analyze the word order in the given source sentence.
The format of CoT is shown in Table~\ref{tab:prompt_cot}.
We notice that GPT-3.5 often fails to identify the modifiers and the modified elements from the given sentence, probably because of a lack of general linguistic knowledge. 
After we hint the modifiers and the modified elements of the given sentences in the prompt, GPT-3.5 can correctly analyze and address their order for 7 out of 10 testing examples.

Our pilot study demonstrates that understanding grammar rules is a complex procedure and still presents significant challenges to current LLMs.
Although several works~\cite{tanzer2024a,geminiteam2024gemini} claim that LLMs are able to \textit{read} grammar books with larger context lengths, it is still questionable whether they truly understand the grammar rules in the books.

\section{Assisting Human Translation} 
Here we discuss the potential of applying \textsc{DiPMT++} to realistic scenarios. 
Through a user study, we show that we can use LLMs to assist humans in understanding an extremely low-resource language, even if both the LLMs and the humans have no prior knowledge about this language.

\subsection{Study Design}
Our  user study aims to investigate to what extent an LLM can help humans translate a completely unseen language.

\paragraph{Settings}
We compare three settings: 
(1) \textbf{LLM Only:} An LLM is prompted with \textsc{DiPMT++} and outputs a translation. 
(2) \textbf{Human Only:} We ask humans to use the given linguistic resources (i.e., a dictionary and a corpus of parallel sentences) to perform the translation.
(3) \textbf{Human + LLM:} We provide humans with an LLM for assistance, in addition to the linguistic resources. Humans can refer to the initial translation results from the LLM.

\paragraph{Data \& Model}
From the \textit{easy} subset of \textsc{ZhuangBench}, we sample 20 instances for zh2za and 40 for za2zh. 
We use Qwen-14B-chat for experiments. This open-source model performs decently on \textsc{ZhuangBench} and has a medium size of parameters with affordable computational cost.

\paragraph{Participants}
We recruit 6 graduate students majoring in NLP, who are native speakers of Chinese but have no prior knowledge of Zhuang. 
The participants are not given training materials regarding the Zhuang language before carrying out the task. We only train them to use the interface through a few demonstrations.
Each testing instance is translated by at least 4 participants. 
For a given testing instance, some participants are asked to translate it with LLM assistance while others are without it.

See details of the study in Appendix~\ref{appendix:user_study}.

\subsection{Results and Analysis}

\begin{table}[t]
\setlength\tabcolsep{3.5pt}
\centering
\begin{small}
\begin{tabular}{l|cc|cc}
\toprule
\multirow{2}{*}{\textbf{Setting}} & \multicolumn{2}{c|}{\textbf{zh2za}} & \multicolumn{2}{c}{\textbf{za2zh}} \\
& \textbf{Time $\downarrow$} & \textbf{Score $\uparrow$} & \textbf{Time $\downarrow$} & \textbf{Score $\uparrow$} \\
\midrule
LLM Only & ~~~~3s & 29.0 / 56.3  & ~~~~3s & 30.3 / 27.9\\
\midrule
Human Only &  309s & 40.2 / 67.6  & 131s & 43.2 / 39.2  \\
Human + LLM & 258s  & 42.7 / 68.4 & 133s & 46.1 / 43.6 \\
\bottomrule
\end{tabular}
\end{small}
\caption{Average time for translating an instance and the translation performance in the user study. The numbers in the Score column are BLEU and chrF. The time of the LLM Only setting is obtained using an A800 GPU.}
\label{tab:user_study}
\end{table}

\paragraph{Quality and Efficiency}
Table~\ref{tab:user_study} shows the study results.
Surprisingly, with the provided linguistic resources, the participants can properly translate simple sentences written in an unseen language into their native language.
Despite costing much time, the average BLEU scores of their translations are 10+ points higher than those of the LLM.

Providing initial translation output from the LLM yields improvement in human translation quality.
For zh2za translation, the LLM helps increase the human performance by 2.5 BLEU while the improvement is 2.9 BLEU for za2zh translation. 

Furthermore, the LLM greatly boosts the efficiency of zh2za translation. The participants save 17\% of their time on average, as
they can leverage the LLM's output rather than crafting translations from scratch. 
For za2zh translation, we observe no obvious difference in terms of efficiency between the two settings. 
It is probably because in the Human Only setting, the participants, who are native Chinese speakers, excel in identifying plausible Chinese words from the given prompt and structuring them into coherent sentences. 
This process requires less time than meticulously verifying the LLM-generated output.

\paragraph{Human Actions}
During the za2zh translation, the participants perform an average of 2.1 dictionary searches and 1.3 corpus searches for each testing instance. 
Conversely, for the harder zh2za translation, we observe more frequent searches: an average of 3.5 dictionary searches and 5.4 corpus searches.

We note that many participants exhibit a pattern of switching between searches conducted in two languages, aiming to find $n$-gram evidence to support the translation of specific words or phrases.  
See examples in Appendix~\ref{appendix:user_study}.
This strategy aligns with the retrieval-augmented generation (RAG) framework, which involves alternating between retrieval and generation stages. Such observations offer insights into innovative solutions for on-the-fly language learning.

\paragraph{Broader Applications}
Besides aiding humans in translation, the LLMs enhanced with the \textsc{DiPMT++} framework have broader applications for low-resource languages. These include education for underrepresented languages, preservation of endangered languages, and research into historical or extinct languages. 
We anticipate that by these techniques, researchers can better contribute to  the linguistic diversity worldwide.

\section{Conclusions}
In this paper, we investigate whether LLMs can learn a completely new language on the fly. Our experiments on the Zhuang language show that LLMs can rapidly grasp an unseen language through proper ICL. 
However, challenges persist in analyzing more intricate morphological phenomena and achieving a more comprehensive and precise understanding of syntax. 
We hope that \textsc{ZhuangBench} and \textsc{DiPMT++} can encourage more research efforts on efficient methods for underrepresented languages.

\section*{Limitations}
\paragraph{Scale of Evaluation}
Due to the limited language resources, the evaluation scale is relatively small. Our evaluation is based on only 200 Zhuang-Chinese testing instances and 50 Kalamang-English testing instances.
We plan to expand the size of the testing set.

\paragraph{Typology of Studied Languages}
Despite belonging to different language families (Krai-Dai and Sino-Tibetan, respectively), Zhuang and Chinese share similarities.
Like Chinese, Zhuang exhibits few inflectional morphologies and has many loanwords from Chinese.
The similarities between Zhuang and Chinese may lead to overly optimistic conclusions.
More research on languages with larger differences, such as that we conduct on Kalamang and English, would provide a more comprehensive understanding.

\paragraph{Scope of Studied Methods}
Our exploration of explicitly learning syntactic information is limited to analyzing a specific syntactic phenomenon with CoT reasoning. 
Other potential methods, such as using external grammar books might may be suitable for more powerful LLMs.
We did not do this due to budget constraints.

\section*{Acknowledgments}
This work is supported in part by NSFC
(62161160339) and Beijing Science and Technology Program (Z231100007423011).
We thank the anonymous reviewers for their valuable suggestions.
We thank Mingxu Tao, Zirui Wu, Zhibin Chen, and Quzhe Huang for their help in this work.
For any correspondence, please contact Yansong Feng.

\bibliography{anthology,custom}

\clearpage

\appendix

\section{The Zhuang Language}
\label{appendix:zhuang}
Zhuang is a group of Kra–Dai languages spoken by the Zhuang people of Southern China in the province of Guangxi and adjacent parts of Yunnan and Guangdong. 
It is used by more than 16 million people. 
The current official writing system for Zhuang is the Latin script. 
Zhuang is considered an isolating language with little inflectional morphology.
In this work, we focus on Standard Zhuang, the official standardized form of the Zhuang language.

Zhuang has been largely overlooked in current NLP research, evidenced by the absence of an open-source corpus dedicated to the language. 
The lack of available training data means that popular open-source multilingual models like mBERT~\cite{devlin-etal-2019-bert}, BLOOM~\cite{workshop2022bloom}, and NLLB~\cite{costa2022no} do not include any support for Zhuang. 
Even competitive commercial models like GPT-3.5 and GPT-4 exhibit near-zero proficiency when it comes to Zhuang. 
This lack of support underscores the challenges faced in developing NLP solutions for low-resource languages like Zhuang.

\section{Dataset Collection and Statistics}
\label{appendix:dataset}

\paragraph{Collection Details}
Here we explain how we collect parallel sentences from the official government reports.
All the versions of the official government reports in other languages strictly correspond to the Chinese version, to ensure consistency in conveying information. The Chinese and Zhuang versions of the official reports have the same number of paragraphs, with each corresponding paragraph containing the same number of sentences. Therefore, we can achieve sentence-level mapping in a fully automated manner by segmenting the reports into sentences and then aligning them in sequence, without the need for manual annotation. 

\paragraph{Data Checking}
We check the data in \textsc{ZhuangBench} and ensure that it contains no information that names or uniquely identifies individual people or offensive content.

\paragraph{Statistic of Parallel Corpus}
In Table~\ref{tab:parallel_dataset}, we report the data statistics of the parallel corpus in \textsc{ZhuangBench}.

\begin{table}[h]
\centering
\begin{small}
\begin{tabular}{lr}
\toprule
Number of Instances & 4,944 \\
Avg. Length in Chinese (by character) & 20.3 \\
Avg. Length in Chinese (by word) &  12.6 \\
Avg. Length in Zhuang (by word) &  12.5 \\
Avg. Depth of Dependency Trees & 2.9 \\
\bottomrule
\end{tabular}
\end{small}
\caption{Data statistics of \textsc{ZhuangBench} parallel corpus. The dependency tree is built in Chinese. The average Chinese word count and dependency tree depth are obtained using spaCy.}
\label{tab:parallel_dataset}
\end{table}

\paragraph{Statistic of Translation Test Set}
In Table~\ref{tab:test_dataset}, we report the data statistics of the translation test set in \textsc{ZhuangBench}. 
We give an example for each level in Table~\ref{tab:level_examples}.

\begin{table}[t]
\setlength\tabcolsep{3pt}
\centering
\begin{small}
\begin{tabular}{l|r|r|r}
\toprule
\textbf{Statistics} & \textit{easy} & \textit{medium} & \textit{hard} \\
\midrule
Number of Instances & 75 & 60 & 65 \\
Avg. Length in Chinese (by character) & 14.2 & 25.5 & 33.8 \\
Avg. Length in Chinese (by word) &  10.4 & 16.9 & 21.8 \\
Avg. Length in Zhuang (by word) &  9.5 & 15.2 & 19.7 \\
Avg. Depth of Dependency Trees & 2.9 & 3.5 & 3.8 \\
\bottomrule
\end{tabular}
\end{small}
\caption{Data statistics of \textsc{ZhuangBench} translation test set. The dependency tree is constructed in Chinese. The average Chinese word count and dependency tree depth are obtained using spaCy.}
\label{tab:test_dataset}
\end{table}

\section{Implementation Details}
\label{appendix:implementation_details}
Here we report the implementation details of \textsc{DiPMT++}.

\paragraph{Prompt Construction}
In Table~\ref{tab:prompts} we provide an example of the prompt for Zhuang-to-Chinese translation used in \textsc{DiPMT++}.
We use 3-shot exemplars, as our preliminary experiments show that using three exemplars can achieve decent performance while having affordable computational costs.
For each word in the source sentence, we provide two possible meanings from the dictionary.

\paragraph{Dictionary Expansion} 
For bilingual lexicon deduction, we use the GIZA++ implementation by Giza-py\footnote{\url{https://github.com/sillsdev/giza-py}}. We set the confidence threshold as 0.6.
For synonym expansion, we use an open-source Chinese synonym list~\footnote{\url{https://github.com/jaaack-wang/Chinese-Synonyms}}.

\paragraph{Hyperparameters of Prompting}
For the prompt-based method, we use the greedy search outputs from LLMs, without doing a hyperparameter search.

\paragraph{Hyperparameters of Finetuning}
For mT5, we adopt the seq2seq training script for machine translation by Transformers~\cite{wolf-etal-2020-transformers}. The batch size is 4. The learning rate is 5e-5. We train the model for 3 epochs, evaluate the checkpoints every 500 steps and select the best-performing one.

For Llama-2-7B, we use the DeepSpeed framework~\cite{rasley2020deepspeed}. We use the instruction "Please translate the following Chinese sentence into Zhuang:" in the prompt. The batch size is 64. The learning rate is 2e-5. We train the model for 3 epochs, evaluate the checkpoints after each epoch and select the best-performing one.

\section{Example Output}
\label{appendix:case_study}
Here we provide the sample output from different methods. 
Table~\ref{tab:case_study_za2zh} shows an example of Zhuang-to-Chinese translation. Table~\ref{tab:case_study_zh2za} shows an example of Chinese-to-Zhuang translation.

\section{Experiments on \textsc{MTOB}}
\label{appendix:kalamang}
Here we report the details for the experiment on \textsc{MTOB}~\cite{tanzer2024a}. 

We choose the \textbf{W} + \textbf{S} setting in the original paper, which uses a dictionary and a corpus of parallel sentences, similar to the setting of \textsc{DiPMT++}.
We follow the original data split, using 50 testing instances for English-to-Kalamang translation and 50 testing instances for Kalamang-to-English translation.

In terms of the implementation of \textsc{DiPMT++}, we run GIZA++ on the parallel corpus and add 412 English words and 345 Kalamang words to the dictionary. For synonym expansion, we use an open-source synonym list extracted from WordNet\footnote{\url{https://github.com/zaibacu/thesaurus}}.

Since the dictionary and parallel corpus in \textsc{MTOB} are only 10\% of those in \textsc{ZhuangBench}, the general translation scores on Kalamng are much lower than those on Zhuang.

\textsc{MTOB} is released under the MIT license.  Our use of this artifact is consistent with its intended use.

\section{Experiments on Seen Languages}
\label{appendix:seen_lang}
We also evaluate \textsc{DiPMT++} on more low-resource languages that might have been seen during pre-training, i.e., the languages that LLMs can translate with a non-zero BLEU score by zero-shot prompting. 
In this way, we could show a clear picture of the applicability of our method.

\paragraph{Languages} We use 7 low-resource languages including Estonian (et), Lithuanian (lt), Latvian (lv), Macedonian (mk), Slovak (sk), Albanian (sq), and Filipino (tl). 

\paragraph{Setup} For each language, we use the 2,009 publicly available sentences from Flores-200~\cite{nllb-22}. We sample 200 sentences from it for testing and the rest form the corpus for retrieval. We use the dictionaries from MUSE~\cite{lample2018word}, consisting of 5K entries. We use Llama-2-13B-chat as the backbone model. 

\paragraph{Results}
We report the results in Table~\ref{tab:seen_language}.
The backbone model already has varied abilities regarding the studied languages, achieving non-zero performance through zero-shot direct prompting. 
This confirms that the model might already see texts in these languages during pretraining.
Regarding the very low-resource languages, of which the model's ability is weak (BLEU < 10), such as Estonian, Lithuanian, Latvian, and Albanian, \textsc{DiPMT++} outperforms DiPMT and direct prompting by a large margin.
In terms of the mid-resource languages, of which the model has a decent ability (BLEU > 20), such as Afrikaans, Indonesian, and Malay, the advantage of \textsc{DiPMT++} is less pronounced.
In conclusion, our method also works on low-resource languages already seen by LLMs, especially on the extremely low-resource ones, of which the model's original ability is poor.

\begin{table*}[t]
\setlength\tabcolsep{2.8pt}
\centering
\begin{small}
\begin{tabular}{l|cc|cc|cc|cc|cc|cc|cc}
\toprule 
\multirow{2}{*}{\textbf{Method}}  & \multicolumn{2}{c|}{\textbf{et}} & \multicolumn{2}{c|}{\textbf{lt}} & \multicolumn{2}{c|}{\textbf{lv}} & \multicolumn{2}{c|}{\textbf{mk}} & \multicolumn{2}{c|}{\textbf{sk}}  & \multicolumn{2}{c|}{\textbf{sq}}   & \multicolumn{2}{c|}{\textbf{tl}} \\
 & \textbf{en2et} & \textbf{et2en} & \textbf{en2lt} & \textbf{lt2en} & \textbf{en2lv} & \textbf{lv2en} & \textbf{en2mk} & \textbf{mk2en} & \textbf{en2sk} & \textbf{sk2en} & \textbf{en2sq} & \textbf{sq2en} & \textbf{en2tl} & \textbf{tl2en} \\
 \midrule
Direct & 5.3 & 12.0 & 4.5 & 10.6 & 4.2 & 10.4 & 11.3 & 31.4 & 12.2 & 32.5 & 3.5 & 9.8 & 13.5 & 27.8 \\
\textsc{DiPMT} & 8.1 & 16.6 & 6.7 & 15.5 & 7.0 & 17.0 & \textbf{15.0} & \textbf{34.2} & \textbf{14.4} & 34.7 & 5.8 & 15.9 & 16.9 & \textbf{32.6} \\
\textsc{DiPMT++} & \textbf{9.1} & \textbf{18.3} & \textbf{7.7} & \textbf{17.0} & \textbf{8.8} & \textbf{18.2} & 14.6 & \textbf{34.2} & \textbf{14.4} & \textbf{35.4} & \textbf{9.7} & \textbf{17.5} & \textbf{18.3} & 31.5 \\
\bottomrule
\end{tabular}
\end{small}
\caption{Performance of different methods on languages seen by Llama-2-13B-chat. We use 3-shot exemplars for \textsc{DiPMT} and \textsc{DiPMT++}. The best scores are made \textbf{bold}.}
\label{tab:seen_language}
\end{table*}

\section{Learning Syntax Explicitly}
\label{appendix:learn_explicitly}
We conduct a pilot study to investigate whether LLMs can comprehend the the syntactical information explicitly given in the prompt.
We focus on the order of modifiers and modified elements, which differs greatly between Zhuang and Chinese.
We select 10 zh2za testing samples where GPT-3.5 fails on the order of modifiers and modified elements, and try different strategies to incorporate this syntactical rule into the prompt. 

First, we attempt to directly declare a rule of word order between modifiers and modified elements in the prompt, i.e., \textit{different from Chinese, Zhuang puts the modifiers after the modified elements}.
GPT-3.5 correctly addresses only 1 of the 10 testing examples.

Second, we use chain-of-thought (CoT) reasoning to encourage the model to analyze the word order in the given source sentence.
The format of CoT is shown in Table~\ref{tab:prompt_cot}.
We notice that GPT-3.5 often fails to identify the modifiers and the modified elements from the given sentence, probably because of a lack of general linguistic knowledge. 
After we hint the modifiers and the modified elements of the given sentences in the prompt, GPT-3.5 can correctly analyze and address their order for 7 out of 10 testing examples. 
This shows the potential of explicitly teaching LLMs syntactic rules through CoT.

\section{User Study}
\label{appendix:user_study}

\paragraph{Data}
From the \textit{easy} subset of \textsc{ZhuangBench}, we sample 20 instances for zh2za and 40 instances for za2zh. 
It is more difficult to perform translations from high- to low-resource languages, so we use fewer zh2za instances than za2zh ones to reduce the burden on the participants.

\begin{figure*}[t]
\centering
\includegraphics[scale=0.55]{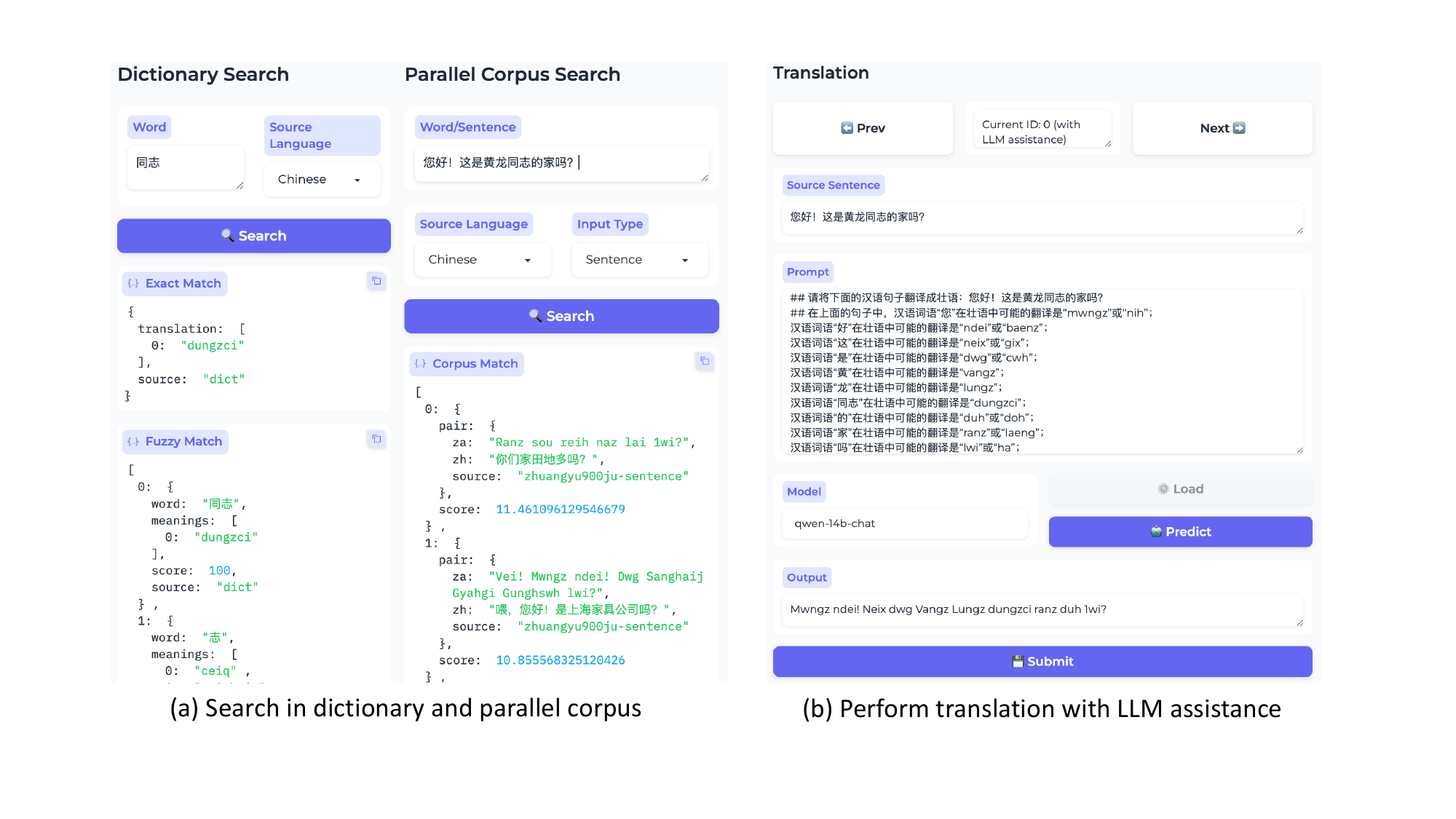}
\caption{Interface for LLM assisted translation.}
\label{fig:interface}
\end{figure*}

\paragraph{Interface}
We develop an interface for the user study, as shown in Figure~\ref{fig:interface}.
To reduce the participants' burden of dictionary searching in the Human Only setting, we also show the prompt generated by \textsc{DiPMT++} to our participants for reference, which already contains possible translations for each word in the sentence to be translated.
The time used for translation and the actions of the users are automatically tracked. 
During the study, the translation instances with and without LLM assistance appear alternately.

\paragraph{Participants}
Before the study, we inform the participants of the scientific use of their translation data and obtain consent from them.
The participants are adequately paid given their demographic.

\paragraph{Human Actions}
In Table~\ref{tab:human_action}, we show an example of how the participant switches between searching Chinese and Zhuang words/phrases during zh2za translation.

\paragraph{Performance Change During the Study}
We do not observe significant performance changes as the participants translate more sentences, since each participant is only assigned 20 instances during the task. Yet, we find that the participants spend slightly more time on the first few instances and their translation speed becomes steady afterwards.

Based on our study, we observe that the benefit of having LLMs utility persists through the translation process. Although humans might become more familiar with the interface, after conducting more translations, it is unlikely for them to accurately remember all the lexical knowledge embedded in these instances, let alone complex syntactical structures. Therefore, LLMs can consistently assist in providing suggestions related to word choice, collocations, and local syntactical structures.

\begin{table*}
\centering
\small
\setlength{\tabcolsep}{5pt}
\begin{tabular}{p{2.0\columnwidth}r}
\toprule
\multicolumn{1}{c}{\textbf{\textit{Easy} Level}} \\
\midrule
\zhsmall{\textbf{Chinese}: 被人骗了，损失了100多万元。} \textit{(I was deceived, and lost more than 1 million yuan.)} \\
\textbf{Zhuang}: Deng vunz yaeuh lo, goem bae bak lai fanh. \\
\midrule
\multicolumn{1}{c}{\textbf{\textit{Medium} Level}} \\
\midrule
\zhsmall{\textbf{Chinese}: 从公共场所回来、捂着嘴巴咳嗽、吃饭之前，我们都要洗手，防止患病。}\textit{(When we come back from public places, cough with our mouths covered, and before eating, we all wash our hands to prevent getting sick.)} \\
\textbf{Zhuang}: Daj giz vunz lai dauqma\zhsmall{、} goemq bak ae\zhsmall{、} yaek gwn haeux, raeuz cungj aeu swiq fwngz, fuengzre baenzbingh. \\
\midrule
\multicolumn{1}{c}{\textbf{\textit{Hard} Level}} \\
\midrule
\zhsmall{\textbf{Chinese}: 我先是住在监狱旁边一个客店里的，初冬已经颇冷，蚊子却还多，后来用被盖了全身，用衣服包了头脸，只留两个鼻孔出气。} \textit{(I first lived in an inn next to the prison, and it was quite cold in the early winter, but there were still many mosquitoes, and then I covered my whole body with a quilt, covered my head and face with clothes, and left only two nostrils to breathe.)} \\
\textbf{Zhuang}: Gou sien youq aen hekdiemq henz genhyuz ndeu, ngamq haeuj seizdoeng gaenq maqhuz nit, hoeng duznyungz lij lai, doeklaeng gou aeu denz goemq daengx ndang, aeu buh duk naj, cij louz song congh ndaeng doeng heiq. \\
\bottomrule
\end{tabular}
\caption{Examples of three difficulty levels in \textsc{ZhuangBench}. The \textit{text in italics} are the English translations of the Chinese sentences. }
\label{tab:level_examples}
\end{table*}

\begin{table*}
\centering
\small
\setlength{\tabcolsep}{5pt}
\begin{tabular}{p{2.0\columnwidth}r}
\toprule
\zhsmall{\# 请仿照样例，参考给出的词汇，将汉语句子翻译成壮语。} \textit{(Please follow the example and refer to the given vocabulary to translate the Chinese sentences into Zhuang.)} \\
\ \\
\zhsmall{\#\# 请将下面的汉语句子翻译成壮语：好。明天你就要回去了，今天晚上我让我妻子弄几个菜，咱们喝两杯。}\textit{(Please translate the following Chinese sentence into Zhuang: OK. You're going back tomorrow. I'll ask my wife to prepare some dishes tonight and we'll have a drink.)} \\

\zhsmall{\#\# 在上面的句子中，汉语词语“好”在壮语对应的词是“ndei”或“baenz”；汉语词语“明天”在壮语对应的词是“ngoenzcog”或“ngoenzbyug”；...} \textit{(In the above sentence, the Chinese word ``good'' corresponds to the Zhuang word ``ndei'' or``baenz''; the Chinese word ``tomorrow'' corresponds to the Zhuang word ``ngoenzcog'' or ``ngoenzbyug'';...)} \\

\zhsmall{\#\# 所以，完整的壮语翻译是：Ndei. Ngoenzcog mwngz couh yaek baema lo, haemhneix gou heuh yah gou loengh geij yiengh byaek, raeuz ndoet song cenj.} \textit{(So, the complete Zhuang translation is: Ndei. Ngoenzcog mwngz couh yaek baema lo, haemhneix gou heuh yah gou loengh geij yiengh byaek, raeuz ndoet song cenj.)} \\
\  \\
\textit{(More exemplars here)} \\
\ \\

\zhsmall{\#\# 请将下面的汉语句子翻译成壮语：现在，农村许多老年人年轻人会使用手机来买卖商品，很方便。}\textit{(Please translate the following Chinese sentence into Zhuang: Nowadays, many elderly and young people in rural areas use mobile phones to buy and sell goods, which is very convenient.)} \\

\zhsmall{\#\# 在上面的句子中，汉语词语“现在”在壮语对应的词是“seizneix”或“neix”；...} \textit{(In the above sentence, the Chinese word ``nowadays'' corresponds to the Zhuang word ``seizneix'' or``neix''; ...)} \\

\zhsmall{\#\# 所以，完整的壮语翻译是：} \textit{(So, the complete Zhuang translation is:)} \\
\bottomrule
\end{tabular}
\caption{Prompt templates used by~\textsc{DiPMT++}. The \textit{text in italics} are the English translations of the Chinese instructions.}
\label{tab:prompts}
\end{table*}

\begin{table*}[h]
\setlength\tabcolsep{3pt}
\centering
\begin{small}
\begin{tabular}{p{0.6\columnwidth}p{1.5\columnwidth}r}
\toprule
\textbf{Input:} & Seizneix, lajmbanj haujlai bouxlaux bouxcoz rox yungh soujgih daeuj gaicawx doxgaiq, fuengbienh lai lo. \\
\midrule
\textbf{Gold:} & \zhsmall{现在，农村许多老年人年轻人会使用手机来买卖商品，很方便。}\textit{(Nowadays, many elderly and young people in rural areas use mobile phones to buy and sell goods, which is very convenient.)} \\
\midrule
\textbf{mT5-large (Finetune):} & \zhsmall{现在，农村贫困人口多，贫困人口多，贫困人口多。} \textit{(Nowadays, there are many poor people in rural areas, there are many poor people, there are many poor people.)} \\
\textbf{Llama-2-7B-chat (Finetune):} & \zhsmall{现在，农村多少少的农车轮到市场上，便利了。} \textit{(Nowadays, more and more farm wheels from rural areas have come to the market, which is convenient.)} \\
\textbf{Llama-2-7B-chat (\textsc{DiPMT++}):} & \zhsmall{现在我们可以通过手机来买卖，方便很多。} \textit{(Now we can buy and sell through mobile phones, which is much more convenient.)} \\
\midrule
\textbf{Qwen-14B-chat (\textsc{DiPMT}):} &  \zhsmall{现在，乡下的父老、青年都知道用东西买卖很方便。} \textit{(Nowadays, the elders and young people in the countryside know that it is very convenient to buy and sell things.)} \\
\textbf{Qwen-14B-chat (\textsc{DiPMT++}):} & \zhsmall{现在，乡下的老人和青年都知道用手机买卖东西很方便。} \textit{(Nowadays, old people and young people in the countryside know that it is convenient to buy and sell things using mobile phones.)} \\
\midrule
\textbf{GPT-4 (Direct):} & \zhsmall{自从，我们村子里来了许多外来人口，村里的面貌变化很大，蓬勃发展起来了。} \textit{(Since then, many outsiders have come to our village. The appearance of the village has changed a lot and it has developed vigorously.)} \\
\textbf{GPT-4 (\textbf{DiPMT++}):} & \zhsmall{现在，农村里许多老人和青年都知道用手机买卖东西了，方便多了。} \textit{(Nowadays, many old people and young people in rural areas know how to use mobile phones to buy and sell things, which is much more convenient.)} \\
\bottomrule
\end{tabular}
\end{small}
\caption{Exmaples of Zhuang-to-Chinese translation from different methods. The \textit{text in italics} are the English translations.}
\label{tab:case_study_za2zh}
\end{table*}

\begin{table*}[h]
\setlength\tabcolsep{3pt}
\centering
\begin{small}
\begin{tabular}{p{0.6\columnwidth}p{1.5\columnwidth}r}
\toprule
\textbf{Input:} & \zhsmall{现在，农村许多老年人年轻人会使用手机来买卖商品，很方便。} \textit{(Nowadays, many elderly and young people in rural areas use mobile phones to buy and sell goods, which is very convenient.)} \\
\midrule
\textbf{Gold:} & Seizneix, lajmbanj haujlai bouxlaux bouxcoz rox yungh soujgih daeuj gaicawx doxgaiq, fuengbienh lai lo.  \\
\midrule
\textbf{mT5-large (Finetune):} & Aen gvaq, aen gvaq ndawbiengz gunghyez gunghyez cungj aeu ndawbiengz cungj aeu yungh ndawbiengz cungj aeu yungh ndawbiengz ndawbiengz daeuj ndaej ndaej ndaej ndaej ndaej ndaej ndaej ndaej ndaej ndaej ndaej ndaej ndaej ndaej ndaej nda \\
\textbf{Llama-2-7B-chat (Finetune):} & Seizneix, lajmbanj haujlai bouxgeq ndaej sawjyungh soujgih daeuj cawx gaiq sang, ndaej fuengbienh.  \\
\textbf{Llama-2-7B-chat (\textsc{DiPMT++}):} & Seizneix hauqgai sanghbinj gaicawx maz daeuj, lai fuengbienh.  \\
\midrule
\textbf{Qwen-14B-chat (\textsc{DiPMT}):} &  Seizneix, lajmbanj haujlai neix, caetneix\zhsmall{年轻人} lox sawjyungh caek maz gaicawx, lai fuengbienh. \\
\textbf{Qwen-14B-chat (\textsc{DiPMT++}):} & Seizneix lajmbanj haujlai bouxgeq mbanjveiz rox yungh soujgih daeuj gaicawx sanghbinj lai fuengbienh. \\
\midrule
\textbf{GPT-4 (Direct):} & Yienvaiz, pwngzloengh bouxlaeuz sw'ciengz niengh yinzhangz daenghhoz gwn lajmbouj raen raemmbawz, ndei ndaw. \\
\textbf{GPT-4 (\textbf{DiPMT++}):} & Seizneix, lajmbanj miz haujlai bouxgeq mbanjveiz rox yungh soujgih daeuj gaicawx sanghbinj, lai bienh. \\
\bottomrule
\end{tabular}
\end{small}
\caption{Exmaples of Chinese-to-Zhuang translation from different methods. The \textit{text in italics} are the English translations.}
\label{tab:case_study_zh2za}
\end{table*}

\begin{table*}
\centering
\small
\setlength{\tabcolsep}{5pt}
\begin{tabular}{p{2.0\columnwidth}r}
\toprule
\zhsmall{\#\# 请将下面的汉语句子翻译成壮语（翻译时请注意，与汉语不同，壮语中的修饰语放在被修饰的中心词之后）：我们村有一条河。}\textit{(Please translate the following Chinese sentence into Zhuang (Please note when translating that, unlike Chinese, the modifier in Zhuang language is placed after the modified elements): There is a river in our village.)} \\

\ 

\zhsmall{\#\# 在上面的句子中，汉语词语“我们”在壮语对应的词是“dou”；汉语词语“村”在壮语对应的词是“mbanj”；...} \textit{(In the above sentence, the Chinese word ``our'' corresponds to the Zhuang word ``dou''; the Chinese word ``village'' corresponds to the Zhuang word ``mbanj'';...)} \\

\ 

\zhsmall{\#\# 那么让我们来一步一步的翻译：} \textit{(So let’s translate it step by step:)} \\

\zhsmall{在该句中，存在修饰语和被修饰语，修饰语是“我们”，被修饰语是“村”，“我们”翻译成“gou”，“村”翻译成“mbanj”。} \textit{(In this sentence, there are modifiers and modified elements. The modifier is ``our'' and the modified element is ``village''. ``Our'' is translated into "gou" and ``village'' is translated into ``mbanj''.)} \\

\zhsmall{在壮语中，修饰语放在被修饰的中心词之后，所以“我们村”翻译成“mbaij dou”。} \textit{(In Zhuang, the modifier is placed after the modified element, so "our village" is translated as "mbaij dou".)} \\

\zhsmall{基于上述分析，该汉语句子的最终壮语翻译是：Mbanj dou miz diuz dah.} \textit{(Based on the above analysis, the final Zhuang translation of this Chinese sentence is: Mbanj dou miz diuz dah.)} \\
\bottomrule
\end{tabular}
\caption{An example of CoT reasoning in the pilot study of learning syntax explicitly. The \textit{text in italics} are the English translations of the Chinese instructions.}
\label{tab:prompt_cot}
\end{table*}

\begin{table*}[t]
\centering
\begin{small}
\begin{tabular}{p{0.7\columnwidth}r}
\toprule
\textbf{Source Sentence:} \zhsmall{这个果园种有多少种果树？}\textit{(How many types of fruit trees are planted in this orchard?)} \\
\midrule
\textbf{Gold Translation:} Aen suenmak neix ndaem miz geijlai cungj gomak?\\
\midrule
\textbf{User Output:} Aen suenmak neix ndaem miz geijlai cungj faexmak? \\ 
\bottomrule
\end{tabular}
\begin{tabular}
{p{0.3\columnwidth}p{0.3\columnwidth}p{0.3\columnwidth}r}
\toprule
\textbf{Action}  & \textbf{Language} & \textbf{Query} \\
\midrule
Word Search & Chinese & \zhsmall{种类} \\
Word Search & Chinese & \zhsmall{种} \\
Corpus Search & Chinese & \zhsmall{多少种} \\
Corpus Search & Chinese & \zhsmall{这个果园} \\
Corpus Search & Chinese & \zhsmall{这个} \\
Word Search & Zhuang & aen \\
Word Search & Zhuang & neix \\
Word Search & Zhuang & ndaem \\
Word Search & Zhuang & cungj \\
\bottomrule
\end{tabular}
\end{small}
\caption{An example of the participant's actions during zh2za translation.}
\label{tab:human_action}
\end{table*}

\end{document}